%% file: main.tex
\pdfoutput=1

\documentclass[11pt]{article}

\usepackage[final]{acl}

\usepackage{times}
\usepackage{latexsym}

\usepackage[T1]{fontenc}

\usepackage[utf8]{inputenc}

\usepackage{microtype}

\usepackage{inconsolata}

\usepackage{graphicx}

\usepackage{tabularx}
\usepackage{booktabs}

\usepackage{listings}
\title{No Culture Left Behind: ArtELingo-28,\\
a Benchmark of WikiArt with Captions in 28 Languages}

\author{{\bf Youssef Mohamed}\textsuperscript{1}\thanks{\parbox [t] {\linewidth} {Corresponding Authors:\{fname.lname\}@kaust.edu.sa}}\quad {\bf Runjia Li }\textsuperscript{2}\quad {\bf Ibrahim Said Ahmad}\textsuperscript{3} \quad {\bf Kilichbek Haydarov }\textsuperscript{1} 
\vspace{1mm} \\
{\bf Philip Torr}\textsuperscript{2} \quad {\bf Kenneth Ward Church }\textsuperscript{3} \quad {\bf Mohamed Elhoseiny}\textsuperscript{1}\textsuperscript{*} 
\vspace{1mm} \\
\textsuperscript{1}KAUST \quad \textsuperscript{2} University of Oxford \quad\textsuperscript{3} Northeastern University
\vspace{3mm} \\
\small{\url{https://www.artelingo.org/}}
}

\begin{document}

\makeatletter
    \let\@oldmaketitle\@maketitle
    \renewcommand{\@maketitle}{\@oldmaketitle
        \myfigure\bigskip}
    \makeatother
    \newcommand\myfigure{%
    \vspace{-4mm}
    \centering
      \includegraphics[width=\linewidth]{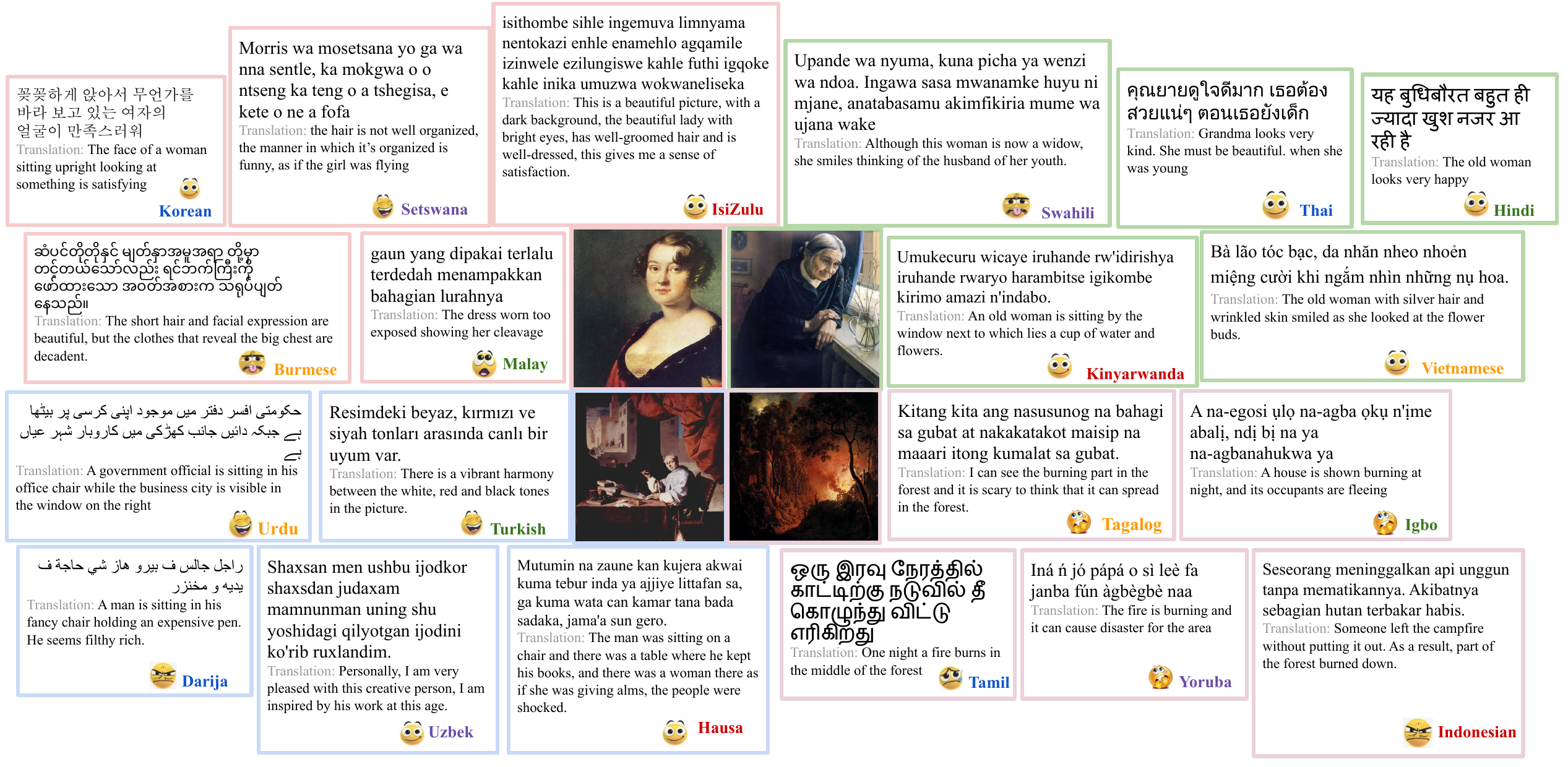}
          \centering
          \refstepcounter{figure}\normalfont{Figure~\thefigure: \textbf{ArtElingo-28 Benchmark:} 
          9 emotion labels with captions in 28 languages.  The $\sim$140 annotations per WikiArt image
          embrace diversity over languages and cultures.
          }

      \label{fig:teaser}
    }

\maketitle

\input{sec/0_abstract}
\input{sec/1_intro}

\input{sec/2_related}
\input{sec/3_dataset}
\input{sec/4_models}

\input{sec/5_experiments}

\input{sec/6_conclusion}

\bibliography{main}

\appendix

\include{sec/X_suppl}

\end{document}

%% file: sec/0_abstract.tex
\begin{abstract}
Research in vision and language has made considerable progress thanks to benchmarks such as COCO.
COCO captions focused on unambiguous facts in English; ArtEmis introduced subjective emotions
and ArtELingo introduced some multilinguality (Chinese and Arabic). However we believe there should be more multilinguality. 
Hence, we present ArtELingo-28, a vision-language benchmark that spans \textbf{28} languages and encompasses approximately \textbf{200,000} annotations (\textbf{140} annotations per image). 
Traditionally, vision research focused on unambiguous class labels, whereas ArtELingo-28 emphasizes diversity of opinions over languages and cultures.
The challenge is to build machine learning systems that assign emotional captions to images. Baseline results will be presented for three novel conditions: Zero-Shot, Few-Shot and One-vs-All Zero-Shot.
We find that cross-lingual transfer is more successful for culturally-related languages.
Data and code will be made publicly available.
\end{abstract}

%% file: sec/1_intro.tex
\section{Introduction}
\label{sec:intro}

A quick review of recent surveys on multimodal AI \cite{cao2023comprehensive, berrios2023towards, zhang2023toward}, reveals just how much the literature is focused on English. The literature on benchmarking \cite{liu2023mmbench, li2023seed} provides an astoundingly similar story. With the pervasiveness of AI technology in our societies, it is essential to make the technology accessible to a wider population.  Although English is widely spoken as a first language or a second language, most of the world (75\% per capita) does not speak English.\footnote{https://www.cochrane.org/news/cochrane-evidence-different-languages}

\autoref{fig:teaser} shows some annotations from ArtELingo-28. For 2000 images from WikiArt, we have $\sim$140 emotion labels per image, as well as captions from annotators with diverse backgrounds covering 28 languages.
Unlike captions in traditional benchmarks such as COCO \cite{lin2014microsoft} and Visual Genome \cite{krishna2017visual} which emphasize unambiguous class labels,
the captions in \autoref{fig:teaser} emphasize subjective opinions over objective facts, and diversity over languages and cultures.
\autoref{fig:teaser} shows 5 annotations from 5 languages for 4 WikiArt images.
Compare, for example, the captions for the first image in Burmese, Malay, Korean and Setswana.
There are differences of opinion in both labels and captions:

\begin{description}
  \setlength{\itemsep}{0pt}
  \setlength{\parskip}{0pt}
 \setlength{\parsep}{0pt}
    \item[emotion labels:] disgust (Burmese), awe (Malay)
    \item[captions:] focus on chest (Burmese \& Malay);\\
    focus on face and hair (Korean \& Setswana)
\end{description}
To advance the field beyond objective facts and unambiguous class labels,
it is critical to embrace diversity and subjective differences of opinion.
Traditionally, vision research has focused on classifying objects in the image in an objective way,
but we prefer to view art as a form of communication between the artist and the audience, where
there is more room for subjectivity and diversity.  Communication depends
on much more than just the pixels in the image such as the cultural backgrounds of
the participants.\footnote{\href{https://www.csuerfsa.org/index.php/news--views/blog/blog/who-created-the-saying-beauty-is-in-the-eye-of-the-beholder}{Blog: Who created the saying that beauty is in the eye of the beholder?}}

To add 25 new languages to ArtELingo-28 required considerable effort.
Amazon Mechanical Turk works well for a few languages, but less so for many of the 25 languages.  
ArtELingo-28 consumed 6.25K hours of work, performed by 220 annotators from 23 countries. Compared to ArtELingo which added just 3 languages, our dataset required significantly more management and coordination; ArtELingo-28 
was managed by a team of 32 coordinators who contributed more than 2.5K hours.

To cover many practical situations, we utilize ArtELingo-28 to build 3 evaluation setups: Zero-Shot, Few-Shot, and One-vs-All Zero-Shot.  The main task evaluates the performance of the generation
of affective explanations. In the Zero-Shot setup, we train a model on a large-scale training dataset in a few high-resource languages. We then evaluate that model on languages that do not appear in the the training data. The Few-Shot setup addresses the situation where we have a few training examples in low-resource languages, in addition to the large-scale training dataset from the Zero-Shot setup. We fine-tune the models from the Zero-Shot setup on the few-shot low-resource data and then evaluate them on the rest of the samples. Finally, in the One-vs-All Zero-Shot setup, we have the large-scale training dataset as well as small-scale data in one language (One). After fine-tuning, we evaluate on the Unseen languages (All). This setup is designed to shed light on pairwise interactions between languages, highlighting cultural effects. 

We observe clusters (cultural groups) forming from our trained models. These groups go beyond writing systems (scripts), capturing cultural connections between languages. 

Additionally, we observe that the multilingual setup is challenging for vision and language models, partly because of the massive vocabulary. We address this challenge by utilizing pretrained multilingual LLMs such as BLOOMZ.

In short, our contributions are:
\begin{itemize}
  \setlength{\parskip}{0pt}
 \setlength{\parsep}{0pt}
    \setlength\itemsep{0.2em}
    \item We collected 200K emotion labels and affective textual explanations in 25 languages on 2000 images (with $\sim$140 annotations/image).
    \item We proposed a benchmark to evaluate the Zero-Shot, Few-Shot, and One-vs-All Zero-Shot performance of Multimodal models. 
    \item We adapt and benchmark four contemporary Vision and Language models to work on our multilingual setup. 
    \item Finally, we study pairwise language transfer revealing insights on cultural differences in emotional perception and expression.
\end{itemize}

%% file: sec/2_related.tex
\section{Related Work}
\label{sec:related}

\textbf{Multimodal Benchmarks: }Benchmarks have always driven the development of many breakthroughs. Imagenet \cite{imagenet} being a perfect example, it led to the development of AlexNet \cite{krizhevsky2012imagenet} which sparked the fire of the deep learning era. Recent benchmarks are moving  towards multimodality. In particular, Vision and Language understanding datasets such as COCO \cite{lin2014microsoft}, Conceptual Captions \cite{sharma2018conceptual}, LAION \cite{schuhmann2022laion}, VQA \cite{Antol_2015_ICCV}, Visual Commonsense Reasoning \cite{Zellers_2019_CVPR}, 
and GQA \cite{Hudson_2019_CVPR} have pushed the frontiers of what is possible. They allowed developing models that can perform complex tasks such as visual grounding, image captioning, text-to-image generation, guided segmentation, and more. Although these datasets are framed as benchmarks to develop vision and language, they mainly cover English language. 

\noindent\textbf{Multilingual Datasets: }However, \emph{``English is NOT a synonym for Natural Language Processing,''} Bender \cite{gradient_benderrule}. 
Cultural background has a great influence on perception. The MARVL dataset \cite{liu2021visually} collected concepts and images that are specified by native speakers with diverse backgrounds. Their dataset revealed a major gap in models trained with English-biased datasets such as Imagenet. In addition to MARVL, other multilingual datasets were proposed such as Multi30K \cite{elliott2016multi30k}, a translated version of Flicker30K \cite{plummer2015flickr30k}, as well as translated versions of COCO \cite{coco_translated,yoshikawa2017stair,li2019coco,al2018automatic}. However, translated datasets have much less diversity compared to natively collected data as shown in the MARVL dataset \cite{liu2021visually}. 

\noindent\textbf{Affective Datasets: }The aforementioned datasets describe the facts and objective reality of the visual input. A recent line of work moved beyond factual captions. ArtEmis \cite{achlioptas2021artemis, mohamed2022okay} collected emotion labels and affective explanations to 80K WikiArt artworks. ArtEmis embellished the facts in images with emotions and commentary. The emotional captions revealed new associations that are ignored in factual captions. The subjective captions in ArtEmis exposed the diversity of human responses. ArtELingo \cite{mohamed2022artelingo} improved the diversity of affective captioning by including Arabic and Chinese. They showed how the different cultures respond differently to the same visual stimulus. In this work, we embrace the cultural differences by introducing 25 more languages.
\autoref{tab:coco_artelingo} compares three datasets; ArtEmis, ArtELingo, ArtELingo-28.

Apart from the Affective Image Captioning line of work, many other emotions-related datasets were introduced. Unimodal datasets that study emotional responses to single input modality such as text \cite{strapparava-mihalcea-2007-semeval, demszky-etal-2020-goemotions, mohammad-etal-2018-semeval, liu2019dens}, image \cite{mohammad2018wikiart,Kosti_2017_CVPR_Workshops}, audio \cite{cowen2019mapping,cowen2020music}, and multimodal \cite{busso2008iemocap}, however, they are all small scale English datasets. 
Emotions shape how humans perceive and process external stimuli, and then act upon those signals. Work in the Psychology literature has explored the effect of cultural background on shaping emotional responses \cite{abu1990romance, henrich2010weirdest, norenzayan2005psychological}. They provide concrete evidence that people from different parts of the world, speaking different languages, perceive the world differently and hence respond differently. ArtELingo-28 is set apart by the inclusion of many languages and hence covering more diverse views of the world.

\noindent\textbf{LLMs: }Recently, Large Language Models (LLMs) have become popular due to a number of major successes, driven in large part by the availability of more data and computational power.
The power of LLMs became apparent with GPT3~\cite{brown2020language}, a major breakthrough over its predecessors. GPT3 can solve unseen tasks in zero-shot settings. Since then, more and more large language models have been developed; Bloom and OPT~\cite{scao2022bloom, zhang2022opt} were developed as open-source alternatives to GPT3; Chinchilla~\cite{hoffmann2022training}, PALM~\cite{chowdhery2022palm}, Megatron-Turing NLG~\cite{smith2022using} are proprietary LLMs with even more parameters (more than 175B parameters in GPT). Notably, LLaMa~\cite{touvron2023llama} is an open-source model with fewer parameters than GPT3. However, it was trained with more than a trillion tokens. Recently, more powerful models have been developed; Llama2/3 \cite{touvron2023llama}, GPT-4 \cite{openai2023gpt4}, and Mistral \cite{jiang2023mistral}.

\noindent\textbf{Multilingual LLMs: }Although the language mentioned above models have achieved impressive results, they mainly focused on English. Multilingual Large Language Models (MLLMs) are a special case of LLMs; unlike LLMs trained in English, MLLMs are pre-trained on text from a more diverse set of languages.  MLLMs are more successful than LLMs on tasks involving cross-lingual transfer.  For cross-lingual transfer, we fine-tune a pre-trained language model on one language and then apply the model at inference time to unseen languages. This approach offers a number of advantages, especially in low-resource scenarios. XLM-R~\cite{conneau2019unsupervised} was one of the first multilingual models to demonstrate cross-lingual transfer capabilities. 
In the space of Large Language Models, many models make use of pretraining data in a variety of languages. However, English tends to dominate as multilinguality is not the main objective.  Bloom~\cite{scao2022bloom} and mT5~\cite{xue2020mt5} are two popular examples of open-source LLM with relatively large contributions of pretraining data from languages beyond English. Building on the success of instruction fine-tuning, Bloomz, and mT0 were introduced \cite{muennighoff2022crosslingual}; these are instruction fine-tuned variants of Bloom and mT5, respectively, with an emphasis on multilinguality.
These models exhibit high-quality cross-language transfer.  This paper utilizes Bloomz to adapt many Vision and Language models to the multilingual setup.

\noindent\textbf{Vision-LLMs: }With LLMs becoming better at generalization, many attempts to integrate modalities were made. For example, in Vision, these models work by injecting visual features into the LLM and then using instruction tuning to teach the LLM reasoning over these features. VisualGPT~\cite{chen2022visualgpt} and Flamingo~\cite{alayrac2022flamingo} are two methods that utilized pre-trained LLM and adapted them to visual features produced from a pre-trained vision encoder model. BLIP2~\cite{li2023blip} proposed using Q-former, a transformer network to map the visual features to the input space of the pre-trained LLM. MiniGPT-4~\cite{zhu2023minigpt} used BLIP2 architecture with a more powerful language model (Vicuna~\cite{chiang2023vicuna}). They curated a high-quality image-text dataset which resulted in big performance gains over BLIP2. LLaVA~\cite{liu2023visual} used image-text instruction following data to align the output of a frozen image encoder with the input of LLaMa. Finally, InstructBLIP~\cite{dai2023instructblip} created an extensive instruction following the image-text dataset using 26 different open-source datasets. 
However, None of these methods studied the cross-lingual capability of such methods in a multilingual setting.
In parallel to adapting LLMs for visual understanding, some works opted for using smaller models and training them from scratch using losses to properly align vision and language modalities. Notably, X$^{2}$-VLM~\cite{zeng2022x} utilized bounding box descriptions to create better vision and language alignment. CCLM \cite{zeng2022cross} added a loss function to align the text from multiple languages. Both models achieved very competitive results on the reported benchmarks.

%% file: sec/3_dataset.tex
\begin{table}[t]
\centering
\scalebox{0.77}{

\begin{tabular}{r|ccc}
    \hline
    & {ArtEmis} & {ArtELingo} & { ArtELingo-28} \\ 
    \hline
    \small{{Image Source}} & WikiArt & WikiArt & WikiArt \\
        \small{{Languages}} &  1 & 3 & 3+ \textbf{25} \\ 
    \small{{\#Images}} & 80k & 80k &  80k(3) , \textbf{2K (25)} \\
       \small{{\#Annotations}} & 0.45M & 1.2M &  1.2M(3) +  \textbf{200K (25)} \\
        \small{{\#Annot/Image}} & 5.68 & 15.3 & 15.3(3) + \textbf{140 (25)} \\
    \small{{Emotions}} & 9 & 9 & 9 (3), \textbf{9 (25) }  \\ 
    \hline
\end{tabular}
}
\caption{\label{tab:coco_artelingo}A Comparison of Datasets. ArtELingo-28 extends ArtELingo \cite{mohamed2022artelingo} with 200K annotations from 25 additional different languages, many are low-resource.}
\end{table}

\section{ArtELingo-28}
\label{sec:dataset}

\autoref{tab:coco_artelingo} compares the difference between the three datasets; ArtEmis, ArtELingo, ArtELingo-28. Our dataset ArtELingo-28 expands horizontally by adding more 25 languages. The challenges for such an expansion are unique. We detail in this section our collection effort with a team of 220 native annotators spanning 23 countries.

We utilize the data collection interfaces from ArtELingo \cite{mohamed2022artelingo}. We ask annotators to carefully examine the image before selecting the most dominant emotion out of 8 emotions.\footnote{Positive: Contentment, Awe, Excitement, Amusement \\Negative: Sadness, Anger, Fear, Disgust} 
In addition, the annotators have the option to select "Something else" if their feelings do not align with any of the 8 emotions. Finally, the annotators are asked to write an explanation of why the image made them feel the selected emotion. Similar to ArtELingo we \textit{aim} to collect annotations from five different annotators for each image. In total, we cover 2000 images. We make sure to have a representative sample of images covering many genres and styles. 
(Please see the full list of art styles and genres in appendix \ref{sec:art_styles}.)
To embrace the different cultural perspectives, we collect annotations for 25 languages from geographically diverse regions. We cover languages from (we include ArtELingo data for completeness):\footnote{Please note that some languages are spoken in multiple regions.}
\begin{itemize}
  \setlength{\parskip}{0pt}
 \setlength{\parsep}{0pt}
    \setlength\itemsep{0.2em} 
    \item \textbf{Africa:} Kinyarwanda, Swahili, IsiZulu, Setswana, Yoruba, Hausa, Igbo, IsiNdebele, IsiXhosa, Emakhuwa.
    \item \textbf{Southeast Asia:} Vietnamese, Indonesian, Thai, Burmese, Malay.
    \item \textbf{Sub-Indian continent:} Tagalog, Tamil, Hindi, Urdu.
    \item \textbf{East Asia:} Korean, \textit{Chinese}.
    \item \textbf{Middle-East:} Turkish, Darija, \textit{Arabic}.
    \item \textbf{Central Asia:} Uzbek, Kazakh, Kyrgyz
    \item \textbf{Europe and North America:} \textit{English}.
\end{itemize}

\noindent\emph{\textbf{Quality Control.}} We deliver training for our annotators. In their native language, we explain the task and the criteria of good explanations; describing image details and relating those details to the selected emotions. The training includes a clear definition of positive emotion labels since not all cultures agree on the meaning of those labels. We provide the detailed descriptions of each emotion in appendix \ref{sec:emo_description}.  

In addition, we employ automatic checks that detect duplicates and incorrect language captions. Finally, we hire native speakers as Language Coordinators to perform manual reviewing of the submitted annotations to guarantee high quality. In addition, Language Coordinators are our point of contact with the annotators, they helped us translate the instructions and training content. Appendix \ref{sec:quality_control} provides more details and statistics about quality control.

\begin{figure}[t]
    \centering

        \scalebox{1.0}{
            \includegraphics[width=1.0\linewidth, height=0.7\linewidth]{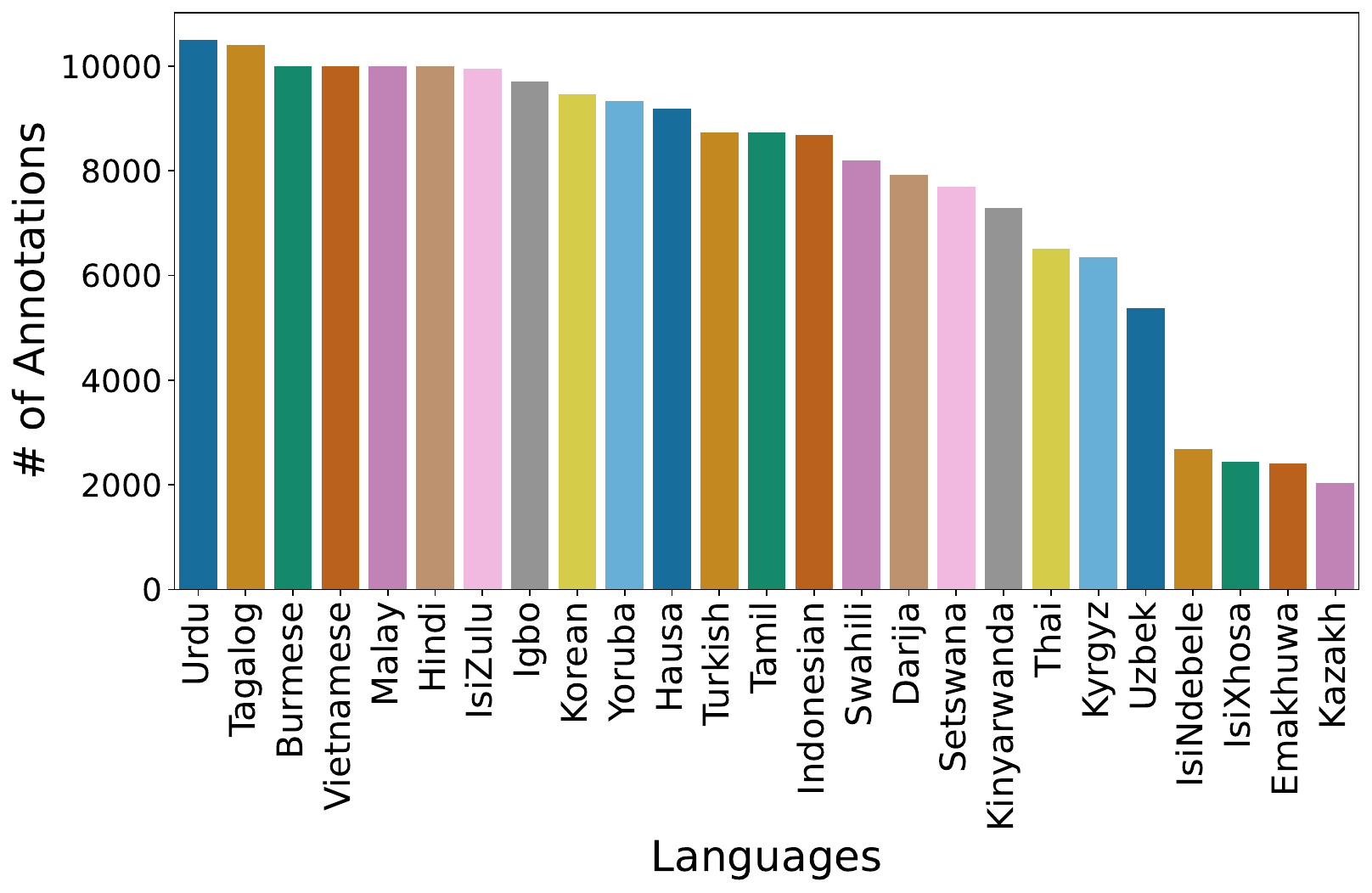} 
            }
        \caption{Number of Annotations per Language}
        \label{fig:num_annotations}
\end{figure}

\subsection{Quantitative Analysis}

\noindent\emph{\textbf{Number of Annotations.}} 
\autoref{fig:num_annotations} reports the number of annotations per language, ranging from 10,493 for Urdo to 2032 
for Kazakh.

\begin{figure}[t]
    \centering
    \includegraphics[width=1\linewidth]{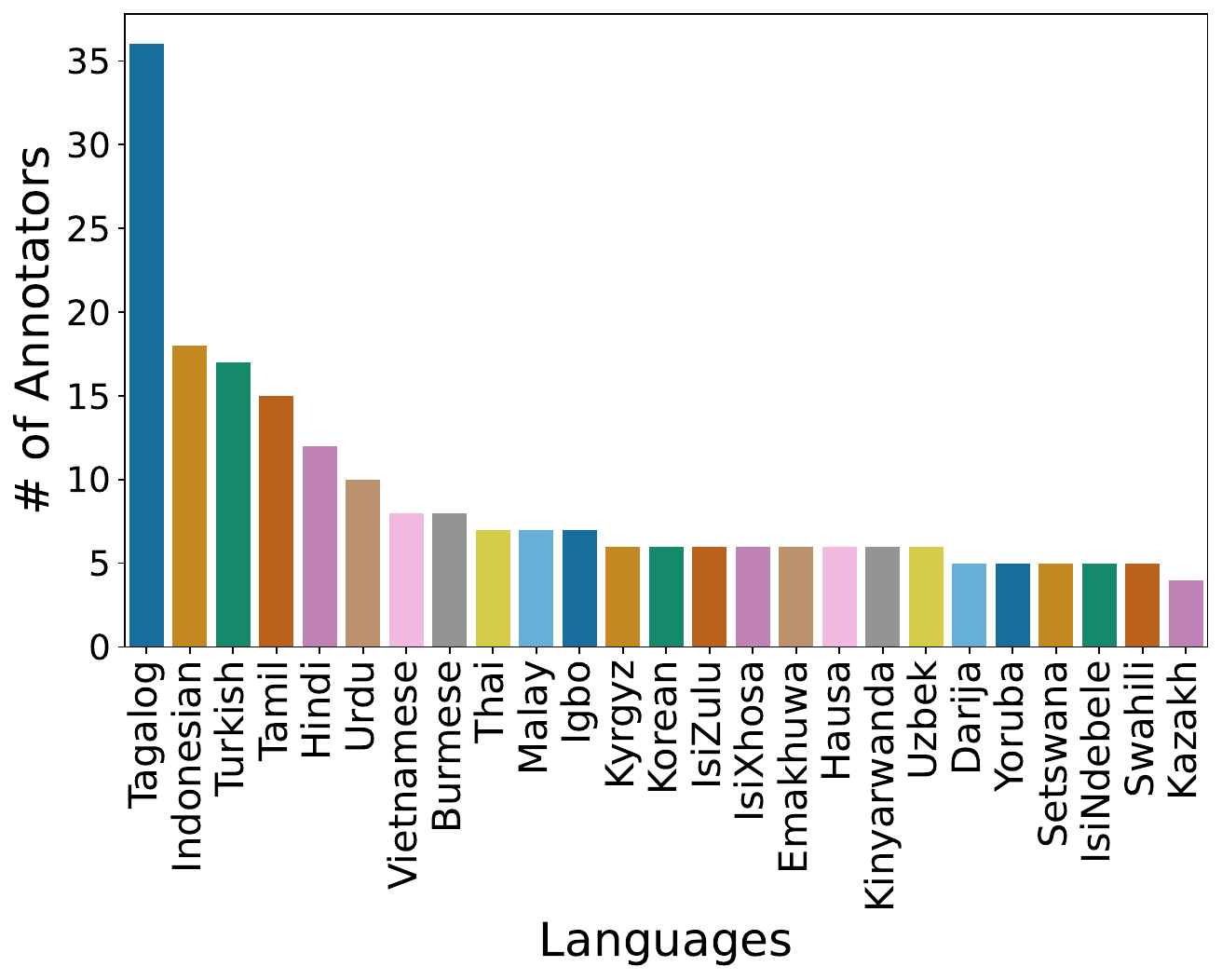}
    \caption{Number of Annotators per Language}
    \label{fig:annotators}
\end{figure}


\noindent\emph{\textbf{Annotators.}}  \autoref{fig:annotators} reports the number of annotators per language. 
A major challenge for ArtELingo-28 was obtaining access to native speakers, especially for low-resource languages. 
Amazon Mechanical Turk was used to collect data, though not to find annotators.
We recruited annotators for Hindi, Urdu, Burmese, Thai, Malay, Vietnamese, Indonesian, Tagalog, Tamil, and Turkish through aiXplain.\footnote{https://aixplain.com/}
For other langauges, we used personal networks to find annotators. 
Although ArtELingo had more annotations, the number of annotators was also proportionally larger making data collection much simpler.  For each language, ArtELingo consumed an average of 10.5K hours/language performed by $\sim$2500 annotators, corresponding to \textbf{4.2 hours/annotator}. In ArtELingo-28, we added 25 languages with an average of 250  hours/language and 8.8 annotators corresponding to \textbf{28.15 hours/annotator}, seven times as much work per annotator. 
These calculations do not include management and coordination efforts.
In total, annotations consumed \textbf{6.25K} hours, plus  \textbf{$\sim$2.5K} hours
for management hours.

\begin{figure}[t]
    \centering
        \includegraphics[width=1.0\linewidth]{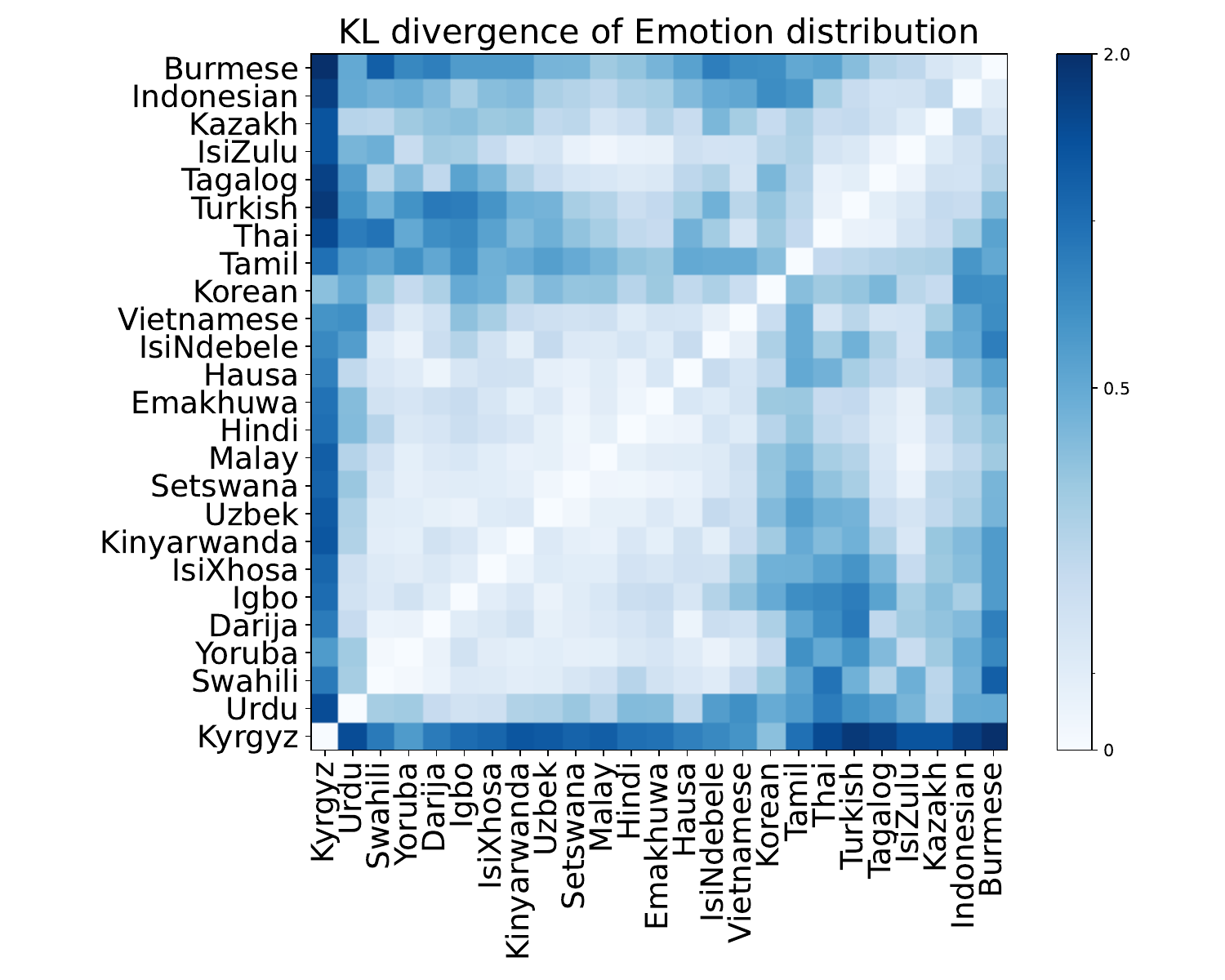} 
    \caption{Kullback-Leibler Divergence between the pairwise emotion distribution. The lighter the color the more emotion agreement between the languages.}
    \label{fig:kl_plot}
\end{figure}

\noindent\emph{\textbf{Emotion Distribution.}} \autoref{fig:kl_plot} shows KL
divergence in emotion labels by language pair.
That is, for a pair of languages $(l_1, l_2 )$ with emotion distributions $(p, q)$, we calculate the emotional disagreement as $D = \sum_{k\in emotions} p_k * \log\frac{p_k}{q_k}$.
We can interpret $D$ as disagreements.
The lighter the color, the more similar the language pair. 
We applied hierarchical clustering to sort languages by agreement in \autoref{fig:kl_plot}.
There are two large clusters in the plot;
the larger cluster contains mostly languages from Africa and the smaller
cluster contains mostly languages from Asia.

%% file: sec/4_models.tex
\section{Models}
\label{sec:models}

We are interested in models able to perform vision and language understanding in many languages. Unfortunately, most open-source state-of-the-art models are heavily biased toward English. Hence, we adapt SOTA general vision and language models to a multilingual setting. The drastic increase in the vocabulary size make such adaptation a challenging task. BLOOMZ tokenizer has a vocabulary size of 250680 tokens compared to only 32000 for Llama2. Hence, the embedding layer of BLOOMZ is much bigger making it harder to align visual features with the language embedding space.

\subsection{LLM-based methods}
We replace the Large Language Model (LLM) with BLOOMZ and introduce a multilingual instruction-following fine-tuning task, resulting in enhanced performance compared to baseline models. This approach is applied to models such as InstructBLIP \cite{dai2023instructblip}, ClipCap \cite{mokady2021clipcap}, MBlip \cite{geigle2023mblip}, and MiniGPT-4 \cite{zhu2023minigpt}.

\noindent\textbf{\emph{Instructions.}}
We utilize a two-stage training process. The first stage aims to align the visual features with the language model input. In this stage, we utilize large-scale datasets, in particular LAION~\cite{schuhmann2021laion} and Conceptual Captions~\cite{sharma2018conceptual}, both have English captions only. In addition, we utilize LAION-2B-multi~\cite{schuhmann2022laion} which is multilingual. We follow MiniGPT-4 \cite{zhu2023minigpt} and use the following instructions during the training,

\vspace{0.2cm}
\noindent\textit{\#\#\#Human: \textless Img\textgreater\textless ImageHere\textgreater\textless /Img\textgreater Could you describe the contents of this image for me? \#\#\#Assistant: }
\vspace{0.2cm}

\noindent The image embeddings replace the \textit{\textless ImageHere\textgreater} tag. As for LAION-2B-multi, we add "\textit{Use only \textless language\textgreater  characters.}" before "\textit{\#\#\#Assistant: }". We use the ISO 639-1 standard for naming languages. 

In the second stage, we use ArtELingo. To ensure better cross-lingual alignment, we group ArtELingo with image IDs. Then, we randomly sample two languages for the same image. We create instructions in the following format,

\vspace{0.2cm}
\noindent\textit{\#\#\#Human: \textless Img\textgreater\textless ImageHere\textgreater\textless /Img\textgreater> Could you describe the contents of this image for me? Use \textless language1\textgreater and \textless language2\textgreater words to describe the image. \#\#\#Assistant: }
\vspace{0.2cm}

We replace \textit{\textless language1\textgreater} and \textit{\textless language2\textgreater} with the two languages we sampled from the dataset. Next, we format the output to be,

\vspace{0.2cm}
\noindent\textit{\textless language1\textgreater:\textless cap1\textgreater. \textless language1\textgreater:\textless cap2\textgreater}
\vspace{0.2cm}

We replace \textit{\textless cap1\textgreater} and \textit{\textless cap2\textgreater} with the captions corresponding to the languages. 
We found this setup to improve the model's alignment across different languages, and improve generalizations to new languages. 

\subsection{Non LLM-based methods}
Additionally, we adapt models that are not based on LLMs by replacing the language encoder with a multilingual language encoder (XLM-R\cite{conneau2019unsupervised}). CCLM provides pre-trained models with the XLM-R backbone, however, their model does not natively support caption generation since it is an encoder-style model. We follow the standard procedure similar to UniLM \cite{dong2019unified} and X$^2$-VLM \cite{zeng2022x} to generate captions via using the [mask] token autoregressively. 
In particular, we start with an empty sentence and append [mask] as the first token. The model then predicts a probability distribution over the vocabulary. We sample the first token from that distribution. Next, we append the [mask] token as a second token and repeat the whole process. We continue the generation of tokens until we reach the maximum sentence length or the end of sentence [eos] predicted by the model.

%% file: sec/5_experiments.tex
\section{Experiments} 
\label{sec:experiments}
This section provides 
baselines for the task of Multilingual Affective Image Captioning. This task
takes three inputs: \emph{image, emotion, language}.  The model is trained to generate an affective caption in the desired language explaining why the image evokes the desired emotion.
The next three subsections discuss 
three experimental setups to evaluate baseline models over a variety of practical situations.
In all three setups, we report standard metrics; BLEU-4 \cite{Papineni02bleu}, METEOR \cite{banarjee2005}, ROUGE \cite{lin-2004-rouge} and CIDEr \cite{vedantam2015cider}. We report summary scores averaged over evaluation languages;
due to page limitations,
detailed results by language are reported in appendix \ref{sec:lang_result}.

Finally, we provide results for emotion label prediction in section \ref{sec:emotion_experiment}.
\begin{table}[t]
    \centering
    \scalebox{0.8}{
        \begin{tabular}{l|cccc}
            \toprule
            Models & BLEU-4 & METEOR & ROUGE & CIDEr\\
            \hline
            InstructBlip & 0.54 & 3.03 & 5.83 & 0.05 \\
            ClipCap & 0.18 & 0.29 &0.23 & 0.02 \\
            mBlip  & 0.23 & 0.32 & 0.73 & 0.04 \\
            MiniGPT-4 & \textbf{1.09} & \textbf{5.8} & \textbf{8.47} & \textbf{0.33}\\
            CCLM & 0.01 & 0.05 & 0.02 & 0.00 \\
            \hline
            \bottomrule
        \end{tabular}
    }
    \caption{\textbf{Zero-shot Performance.} MiniGPT-4 is the best performing model.}
    \label{table:zeroshot}
\end{table}

\subsection{Setup 1: Zero-Shot}
This setup
is intended for cases where we have just a few high-resource
languages with large training sets.
Specifically, 
we consider Arabic, Chinese, and English as high-resource languages.
In the Zero-Shot setting, the system is trained on ArtELingo data \cite{mohamed2022artelingo} in Arabic, Chinese and English,
and tested on 25 other languages in ArtELingo-28.

\noindent\textbf{{Results.}}
\autoref{table:zeroshot} reports the average scores over all the languages. It is evident that MiniGPT-4 \cite{zhu2023minigpt} is the best performing model in this setting, followed by InstructBlip. This aligns with results from LVLM-eHub benchmark \cite{xu2023lvlm} where MiniGPT-4 shows superior performance on open-world scenarios compared to InstructBlip which heavily overfits existing tasks.
However, both models are superior to competition due to their superior pretraining strategies.

ClipCap has a similar architecture to MiniGPT-4 and InstructBlip but it does not undergo a similar pretraining. The results show the importance of high-quality pretraining in improving the model's cross-lingual Zero-Shot performance.

Although the CCLM model is reported to achieve SOTA results when trained on a given task \cite{zeng2022cross}, it suffers greatly when it comes to cross-lingual Zero-Shot performance. CCLM is not based on large language models which limits its reasoning, instruction-following, and generalization abilities.

\begin{figure}[t]
    \centering
        \includegraphics[width=1.0\linewidth]{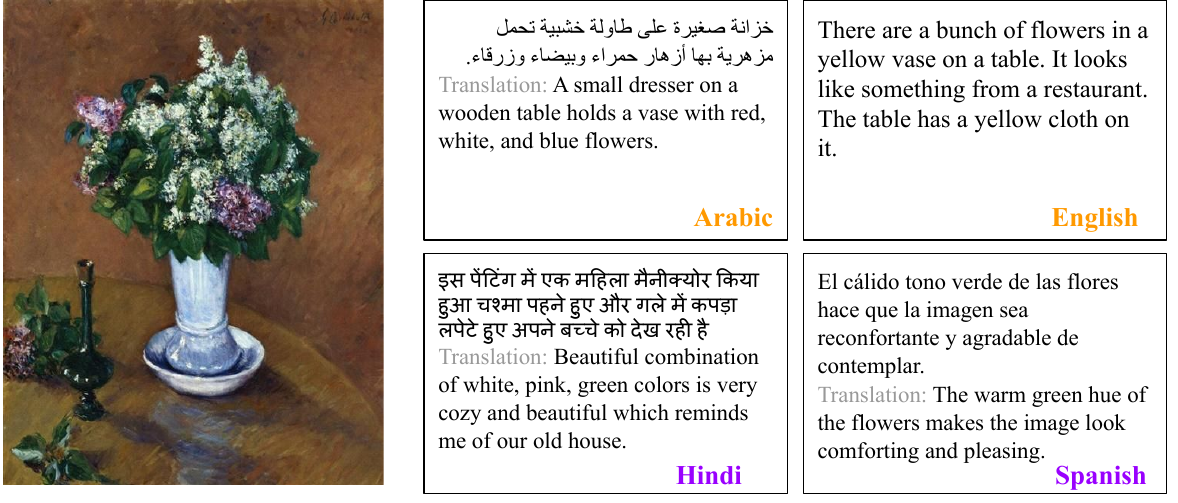}
    \caption{\textbf{Example Zero-Shot Generations.} The top row is the performance on the test data from ArtELingo where the model has seen the languages during training. The second row corresponds to languages that the model has not seen during multimodal training.}
    \label{fig:generations}
\end{figure}

\autoref{fig:generations} showcases some generations from the MiniGPT-4 model. The quality of the generations is quite good, even on the bottom row
where the model was not trained on those on those languages.

\subsection{Setup 2: Few-Shot}
This setup corresponds to the scenario
where we have a modest amount of data ($\sim$7K) from low-resource
languages in addition large amounts of data ($\sim$900K) from high-resource languages.
We start from the Zero-Shot datasets and the Zero-Shot  models, and then further fine-tune the model on the 25 additional languages in ArtELingo-28. We vary the ratio of the training samples from 20\% to 100\%. Please note that not all 25 languages have the same number of annotations. We report the results from MiniGPT-4 since it is the best performing model.

\begin{table}[t]
\centering
\scalebox{0.8}{
    \begin{tabular}{l|cccc}
    \toprule
    Percentage & BLEU-4 & METEOR & ROUGE & CIDEr \\
    \hline
    20\% & \textbf{13.5}  & 14.3 & 20.4 & 0.93\\
    60\%  &13.4 &14.2 & 20.4 & 0.93 \\
    
    80\% & 12.9 & 14.5 & 20.7 & \underline{0.95} \\
    
    100\% & 13.1 & \textbf{14.6} & \textbf{20.9} & \underline{0.95} \\
    \hline
    \bottomrule
    \end{tabular}
}
\caption{\textbf{Few-shot Performance.} We observe a significant performance gain on MiniGPT-4 over the Zero-Shot model. MiniGPT-4 is sample efficient and reaches the best performance with few data points due to the reasoning and generalization ability of its LLM.}
\label{table:fewshot}
\end{table}
\noindent\textbf{{Results.}}
We report the results in \autoref{table:fewshot}. We observe a major improvement over the Zero-Shot model's performance. However, we don't observe a significant improvement in the performance as we use more finetuning data. Hence, we recommend collecting more languages over more samples (\textbf{expand horizontally}) if the objective is cross-lingual transfer.

\subsection{Setup 3: One-vs-All Zero-Shot}
\label{sec:seenunseen}
This setup aims to study the pairwise interaction between languages. We fine-tune the Zero-Shot model on one language, and evaluate on all the other languages. We hypothesize that successful cross-lingual transfer is driven by close cultural connections. For example, 
let model $x$ be fine-tuned on Hindi, while model $y$ is fine-tuned on Hausa. We evaluate both models on Urdu. If both models perform \textbf{similarly}, then there is \textbf{no} underlying cultural relationship between Hindi or Hausa with Urdu. However, if either model performs better, then we can assume an extra cultural connection between Urdu and the best performing model's language. 

\begin{figure}[t]
    \centering
    \includegraphics[width=1.0\linewidth]{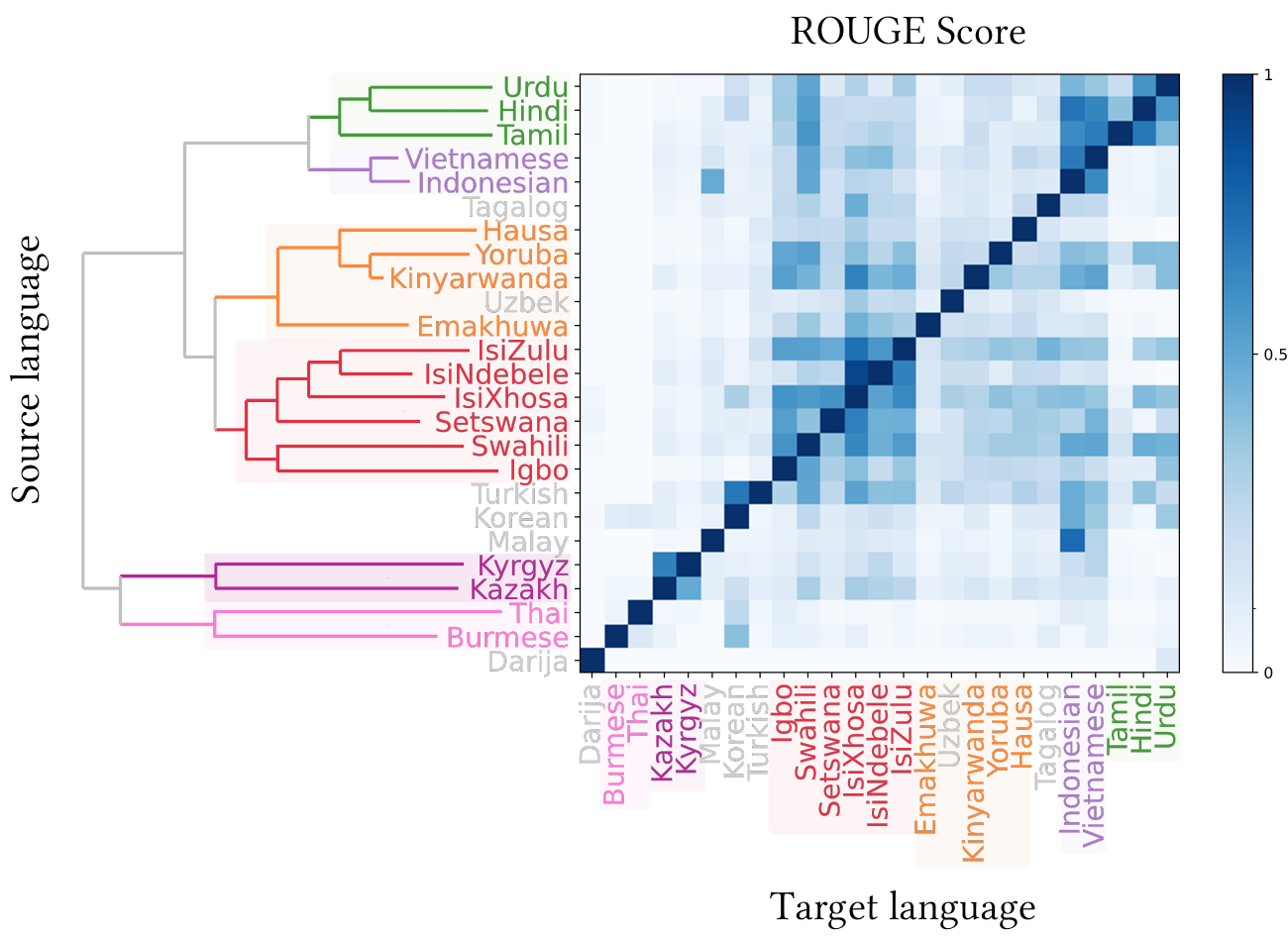}
    \caption{\textbf{One vs All Zero-Shot.} The figure shows the rouge score on the target languages. On the left the clustering reveals cultural connections. The captioning scores reveal groups that align with real world cultural connections. This clustering suggests that our trained models can capture the cultural signal.}
    \label{fig:onevsall}
\end{figure}

\noindent\textbf{{Results.}}
We define the \emph{Source} language as the language used for finetuning the ZeroShot model while the \emph{Target} language is the language being evaluated. 

\autoref{fig:onevsall} shows ROUGE scores by language pair.
Columns are normalized by MinMax scaling.\footnote{\scriptsize sklearn.preprocessing.MinMaxScaler} 
Hierarchical clustering is used to sort languages, so languages with similar scores are grouped near one another. 
Colors were added manually to call out clusters.
By construction, the scores are large along the main diagonal where the same language is
used for both fine-tuning and evaluation.

Interestingly, we see language groupings that reflect cultural connections irrespective of other factors such as writing systems. We can see major clusters representing cultural groups. For example:
\begin{itemize}
  \setlength{\parskip}{0pt}
 \setlength{\parsep}{0pt}
    \setlength\itemsep{0.2em} 
    
    \item \textcolor[HTML]{439d43}{Urdu, Hindi, Tamil:} 
    These languages share considerable history, though
    they use quite different scripts.
    Transfer between these languages is relatively successful compared to other languages.
    \item \textcolor[HTML]{b07bc7}{Indonesian, Vietnamese:} These languages are spoken in Southeast Asia, and are geographically close to one another.
    
    \item \textcolor[HTML]{e22f48}{IsiZulu, IsiNdebele, IsiXhosa, Setswana:} These languages are spoken in Southern Africa. They also belong to a larger cluster of African languages: \textcolor[HTML]{fd8a4c}{Hausa, Yoruba, Kinyarwanda, Emakhuwa, Swahili, Igbo}.
    
    \item \textcolor[HTML]{b8379c}{Kyrgyz, Kazakh:} These two languages are similar to each other. Both Kyrgystan and Kazakhstan share considerable history and tradition. Similarly, \textcolor[HTML]{ff7cd5}{Burmese and Thai}'s cultures are related to one another.
\end{itemize}
We find it interesting to observe such groupings emerging naturally from the data collected from annotators with different backgrounds and traditions. Although some languages share the same writing system characters such as Darija and Urdu, they are very far away in our clustering. This means that the writing system has little to do with this clustering. 

We can see the predominant advantage of collecting data from native speakers. Our trained models are better at embracing the different cultural perspectives. 

\subsection{Emotion Label Prediction}
\label{sec:emotion_experiment}
This task takes a caption as an input and predicts one of the nine emotions as an output. We finetune XLM-roBERTa \cite{conneau2019unsupervised} in multiple settings to show the advantages of finetuning with ArtELingo-28. Our first model called \textit{base} is trained on 900K annotations from ArtELingo consisting of captions in Arabic, Chinese, and English languages. \textit{ArtELingo} model is further finetuned on 200K samples from ArtELingo different from the initial training data. This model simulates collecting more data in high resource languages hoping to positively improve multilingual performance. \textit{ArtELingo-28} model load the \textit{base} model and then finetunes on our ArtELingo-28 dataset. This model corresponds to collecting native multilingual data. Finally, \textit{$ArtELingo$-$28_O$} finetunes XLM-roBERTa only on ArtELingo-28 to measure the usefulness of training using high resource languages before finetuning. In all of our experiments, we finetune the XLM-roBERTa large for 5 epochs. 

Table \ref{tab:emotion_pred} reports the accuracy, precision, recall, and F1 scores for all the models. We evaluate on the test set of ArtELingo-28.

\begin{table}[h!]
\centering
\scalebox{0.77}{
    \begin{tabular}{l|cccc}
        \toprule
        Model & Accuracy & Precision & Recall & F1 \\
        \midrule
        \textit{base} & 0.37 & 0.415 & 0.325 & 0.342 \\
        \textit{ArtELingo} & 0.354 & 0.357 & 0.31 & 0.313 \\
        \textit{$ArtELingo$-$28_O$} & 0.636 & \textbf{0.662} & 0.582 & 0.606 \\
        \textit{ArtELingo-28} & \textbf{0.664} & 0.651 & \textbf{0.628} & \textbf{0.638} \\
        \hline
        \bottomrule
    \end{tabular}
}
\caption{\textbf{Emotion Label Prediction.} Finetuning using ArtELingo-28 is essential to learn the culture specific emotional responses. }
\label{tab:emotion_pred}
\end{table}

\noindent\textit{ArtELingo-28} is the best performing model. Notice the huge gap between \textit{ArtELingo} and \textit{ArtELingo-28}. It reflects the need to collect data from native speakers. Naive scaling of data by collecting more samples from the same languages does not help, in contrast, it seems to harm the performance in our case. Finally, the difference between \textit{ArtELingo-28} and \textit{$ArtELingo$-$28_O$} emphasize the importance of pretraining on a large dataset even if the languages are difference. It shows the ability of the multilingual XLM-roBERTa to do cross-lingual transfer. Appendix \ref{sec:emotion_appendix} include more hyper-parameters and training details.

%% file: sec/6_conclusion.tex
\section{Conclusion}
\label{sec:conclusion}
In summary, ArtELingo-28 addresses a critical gap in evaluating large-scale multilingual Affective Vision and Language understanding. By adding 25 languages and 200K high-quality annotations, including low-resource languages, our dataset embraces the cultural differences. 

Our evaluation setups — Zero-Shot, Few-Shot, and One vs All Zero-Shot — assess affective explanation generation across diverse linguistic contexts. The One vs All Zero-Shot setup extends evaluation to languages beyond the training dataset, revealing cultural connections through cross-lingual transfer performance.

In this work, we introduced a dataset, proposed a benchmark, and adapted four Vision and Language models, overcoming current limitations in multilingual AI evaluation. ArtELingo-28 sets a benchmark for bridging linguistic and cultural gaps in Affective Vision and Language understanding.

\section{Limitations}
\noindent\textbf{Modeling.} Our approach faces limitations in modeling, evaluation, and data. Modeling relies on the quality and availability of multilingual large language models (MLLMs), stressing the need for attention and resources beyond English models. Current evaluation metrics overlook emotional and subjective aspects, necessitating new metrics. To enhance benchmarking, broader language coverage in evaluation datasets is crucial, alongside the collection of more diverse native multilingual data. Addressing these limitations is essential for advancing Multilingual General Purpose Vision Language Models' effectiveness and applicability.

\noindent\textbf{Dataset.} In addition to the limitations of ArtELingo \cite{mohamed2022artelingo}, our dataset reflects the viewpoints of the annotators. We instructed the annotators neither to attack any given group of people based on ethnicity, religion, etc. nor use vulgar language. In addition, We asked our coordinators to reject and report any captions that used vulgar language or attacked a group of people. 

The captions might reflect ideas that might be sensitive in the western cultures. However, they reflect the world views of people from different cultures. We should expect some things and topics to be more or less sensitive depending on culture: dress, gender, religion, drinking, sex, respect for animals, respect for the environment, etc. It is not helpful to demand that all cultures conform to a specific world view.

Our annotators represent a group of people who have internet access and are educated. Most of them speak English in addition to their native language. We leave extending our dataset to include other groups of people to future work. 

Our goal in ArtELingo-28 is to embrace diversity and capture the diverse perceptions of the world held by different cultures. While we may disagree on topics such as religion, dress, and other cultural norms, our position as authors, coming from different cultures and covering three continents, is that it is important to consider these varying perspectives to build culture-aware models, while of-course promoting respect and eliminating, as much as we can, hateful content. We should embrace the fact that what is considered appropriate or inappropriate varies across cultures and try to understand other people. We find it not our job to draw one line for all cultures but to expose this phenomenon that we find worth studying. Our dataset aims to help people understand and respect each other's worldviews, even if disagreement is inevitable on some topics, which may serve as a resource for advancing cultural and cross-cultural psychology.

\section{Acknowledgment}
The authors would like to thank aiXplain for annotators in 10 languages and providing high quality annotations. In addition, we thank all our language coordinators for their amazing effort and support throughout the project. We extend our gratitude to all the annotators for their effort in the data collection. 

This work was supported by King Abdullah University of Science and Technology (KAUST), under Award No. URF/1/5016.

%% file: sec/X_suppl.tex
\onecolumn

\section{Dataset Collection Interface}
\label{sec:interface}
\begin{figure*}[b]
    \centering
    \includegraphics[width=0.8\linewidth]{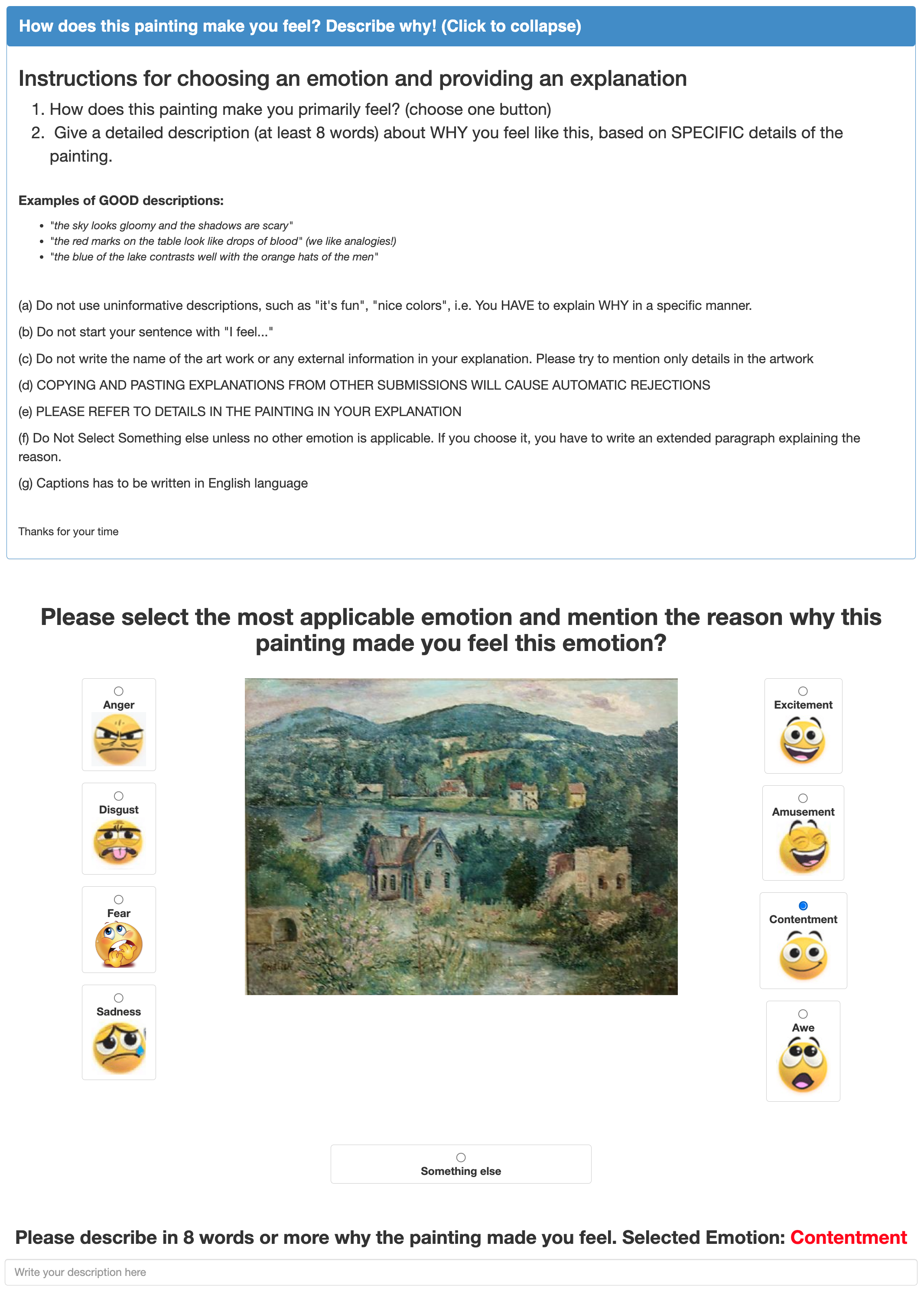}
    \caption{Interface used to collect ArtELingo-23}
    \label{fig:interface}
\end{figure*}

We use the interface in Figure \ref{fig:interface} to collect the annotations for ArtELingo-23. We hire native speakers who are very proficient in English to translate the interface to the respective languages.

\section{Quality Control}
\label{sec:quality_control}
Our quality control includes multiple stages. Below we outline these stages and their details:
\begin{itemize}
    \item \textbf{Hiring Stage: }we have worked with our language coordinators on multiple projects before and they have a proven track record of providing high quality data through their annotators. They know the annotators in person as a result the hired annotators are top quality.
    \item \textbf{Training Stage: }we provide detailed training through a conference meeting for all the annotators and coordinators. We then ask the coordinators to translate the list of instructions and the key points of the training into their respective language to make sure that all annotators understand the instructions completely.
    \item \textbf{Reviewing Stage:}
    \begin{itemize}
        \item We developed automatic scripts that perform duplicate detection, sentiment analysis verification, and language identification. For duplicates we used exact text matching. We utilized a sentiment analysis model fine tuned on ArtELingo (English, Arabic, Chinese) to classify whether the caption is positive or negative. Finally, we utilized the NLLB\cite{costa2022no} language identification fasttext model \footnote{huggingface.co/facebook/fasttext-language-identification} to make sure the languages match our target. Any automatically rejected annotation was marked for the language coordinators to review in detail. Noteworthy, duplicate detection was triggered 20\%, 32\%, 47\% for Thai, Urdu, and Turkish (All languages were supported by AiXplain). We discussed the issue with the annotators; All duplicates were re-annotated and we didn’t observe this issue. The sentiment analysis classifier marked very few instances <1\%. Finally, the language identification also marked <1\% for all languages except Malay and Indonesian since the two languages are very similar to each other. The language identification model classified 75\% of Malay instances as indonesian. However, manual inspection revealed no issues.
        \item The language coordinators manually review the annotations. Overall, the rejection rate in each language was between 1~5\% reflecting the high quality of annotations. The most common mistake encountered was selecting the “something else” emotion label while the caption reflects one of the 8 emotions. For this mistake, we provided extra training that focused on explaining the emotional labels for the annotators. The initial training explained the emotion labels but some languages (Burmese, Turkish, Swahili, Hausa, Indonesian, Korean) required extra training. After that the annotators re-selected the emotional label to align better with our definition of emotions.
        \item Another issue is that some annotators started their sentences with “This image makes me feel …” which heavily influenced the performance of our captioning models. We asked the annotators to fix those annotations by paraphrasing them to more natural sentences. This was mainly encountered in Yoruba, but was fixed early in the annotation process. 
    \end{itemize}
    \item \textbf{Translated Validation:} As a final quality check after the coordinators, we (authors) translated a random sample of 500 annotations from each language to English and performed sanity checks. We didn’t encounter any bad quality annotations.
\end{itemize}

\section{Dataset Analysis}
\subsection{Art Styles and Genres}
\label{sec:art_styles}
In total, we cover a subset of 2000 images from the set of 80K images used in ArtELingo. We make sure to have a representative sample of images covering all the art styles and genres. 

\noindent The art styles are Abstract Expressionism, Action painting, Analytical Cubism, Art Nouveau Modern, Baroque, Color Field Painting, Contemporary Realism, Cubism, Early Renaissance, Expressionism, Fauvism, High Renaissance, Impressionism, Mannerism Late Renaissance, Minimalism, Naive Art Primitivism, New Realism, Northern Renaissance, Pointillism, Pop Art, Post Impressionism, Realism, Rococo, Romanticism, Symbolism, Synthetic Cubism, and Ukiyo-e. 

\noindent While the genres are portrait, landscape, genre painting (misc), religious painting, abstract painting, cityscape, sketch and study, still life, and illustration.
\subsection{Caption Length}
\begin{figure}[t]
    \centering
    \includegraphics[width=1.0\linewidth]{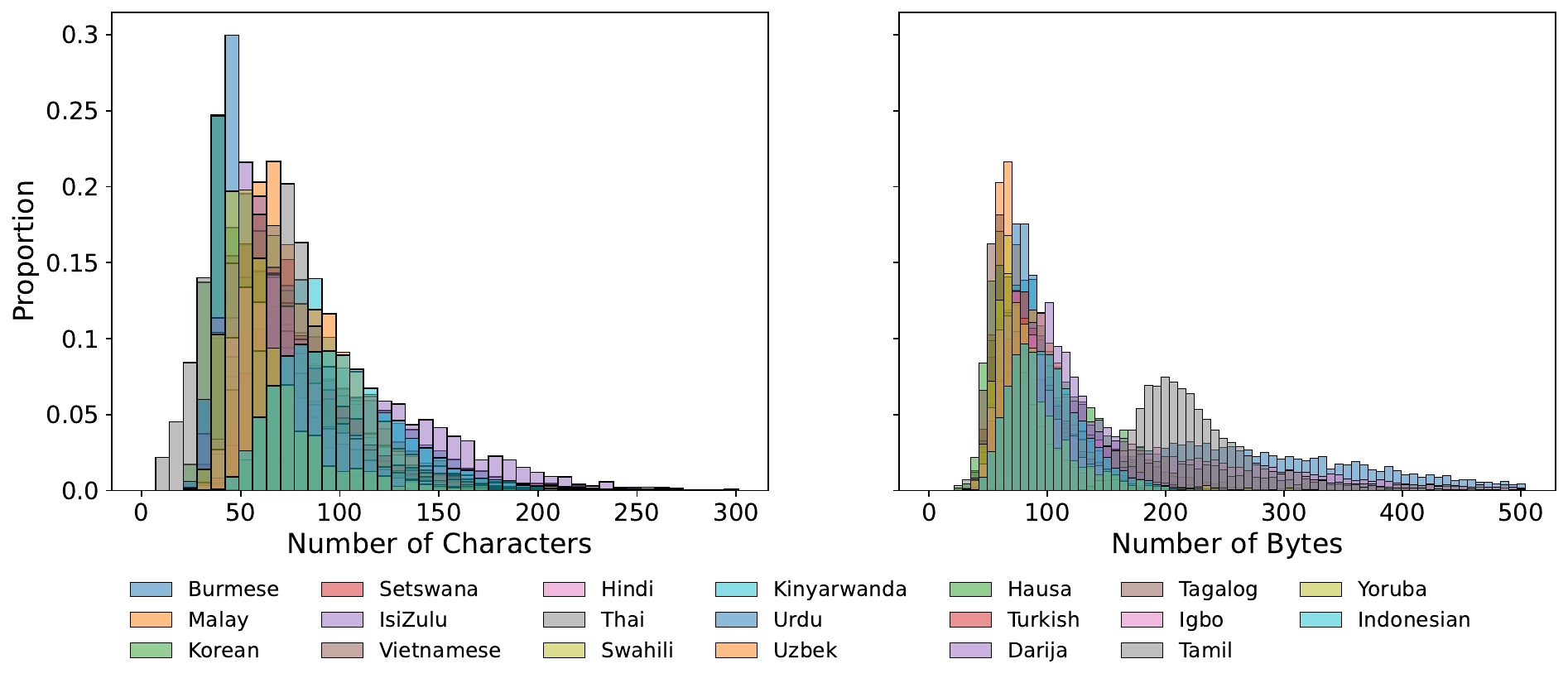}
    \caption{\textbf{Left:} Distribution of the caption length as the number of characters. \textbf{Right:} Distribution of the caption length as number of bytes required for storage.}
    \label{fig:cap_len}
\end{figure}

\begin{figure}[t]
    \centering
    \includegraphics[width=0.8\textwidth]{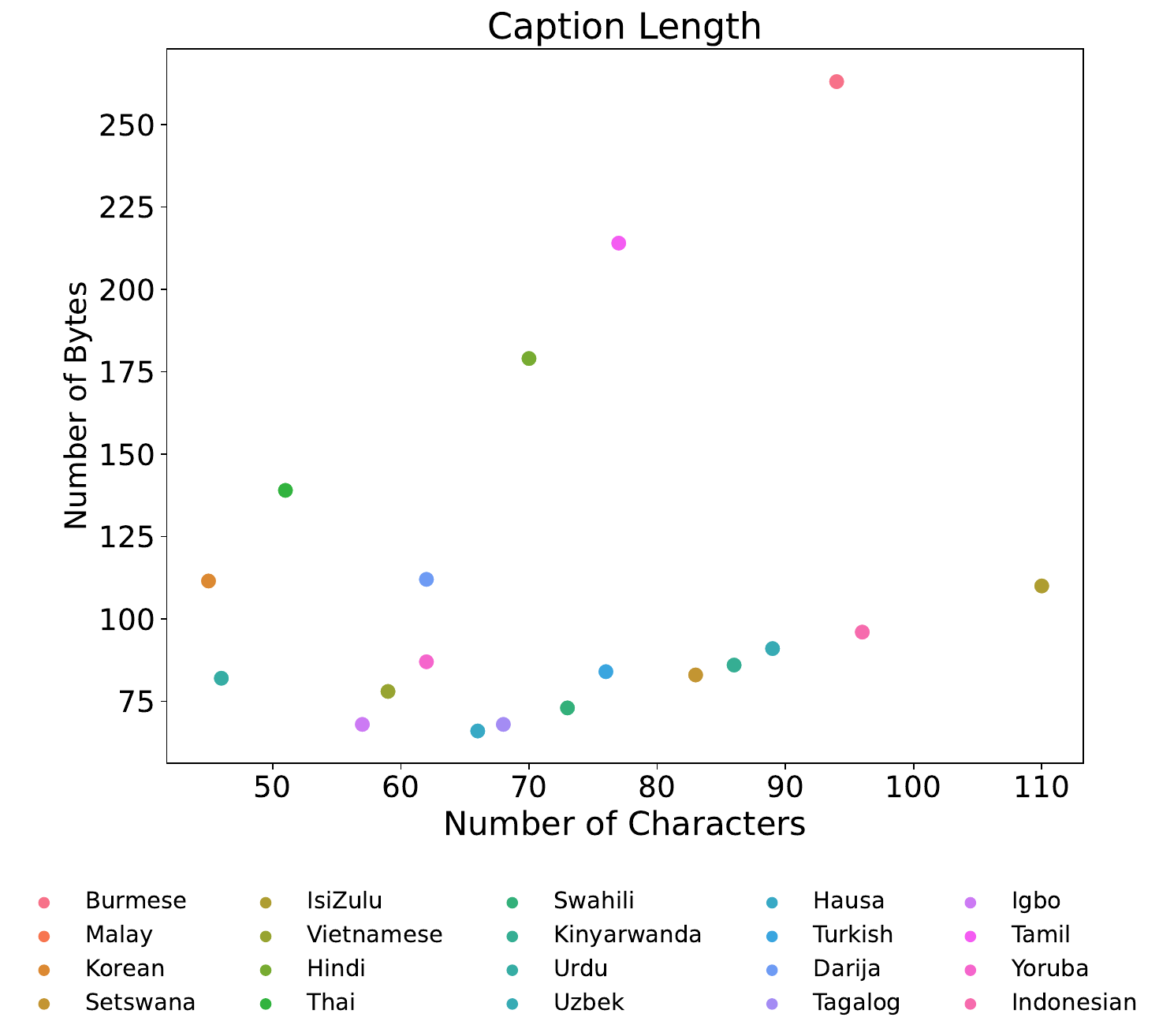}
    \caption{Median Number of Characters vs Number of Bytes per caption}
    \label{fig:len_bytes}
\end{figure}

Figure \ref{fig:cap_len} showed the number of characters and number of bytes per caption in separate plots. We combine both measures in figure \ref{fig:len_bytes} to better understand the effect of character encoding. Most languages have approximately one to one byte to character ratios. While other languages such as Korean and Thai have higher ratios, approximately four and three, respectively. It is interesting to study the effect of this ratio on the reported metrics in the caption generation experiments. Although it is difficult to disentangle the effect of different confounders such as the LLM pretraining, tokenizer, etc.

\subsection{Emotion Distribution}
\label{sec:emo_description}
\begin{figure}
    \centering
    \includegraphics[width=0.9\linewidth]{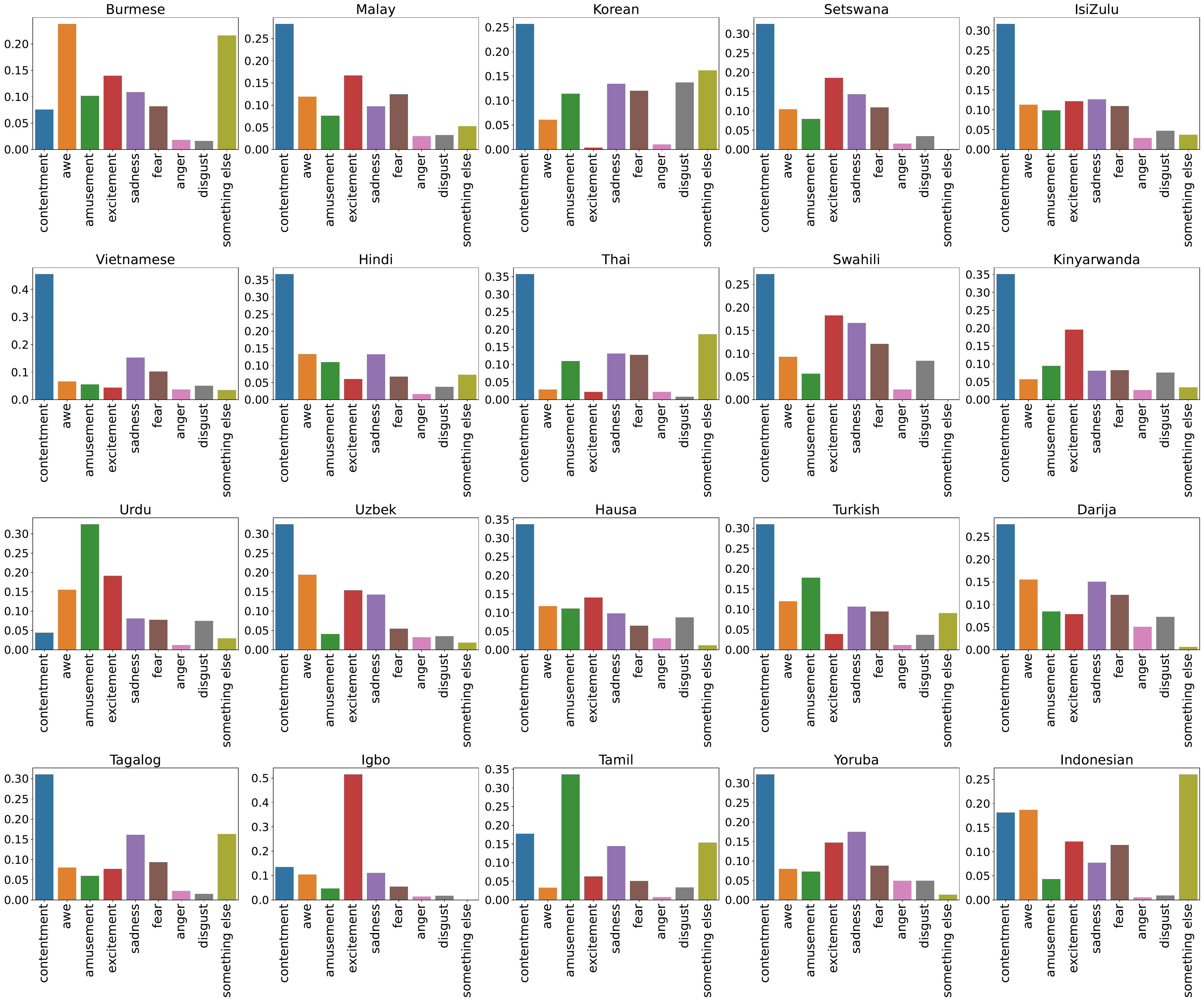}
    \caption{Emotion Distribution for different Languages}
    \label{fig:full_emo}
\end{figure}

Figure \ref{fig:full_emo} shows the emotion distribution for the different languages. Although we provided a detailed description of each emotion label (see Table \ref{tab:emotions}), we still see variations in some languages. 

\begin{table*}[t]
\centering
\begin{tabular}{c|p{0.8\textwidth}}
\toprule
\textbf{Emotion} & \textbf{Description} \\
\hline
Contentment & "A deep sense of satisfaction and inner peace. It often arises when a person feels comfortable, secure, and at ease with their present circumstances.
\textit{Example Caption: A peaceful, sunlit afternoon by the lake, with a person sitting on a comfortable chair, sipping tea, and smiling at the serene view of nature.}" \\
\hline
Excitement & "An intense feeling of enthusiasm, eagerness, and heightened energy. It typically emerges in response to something thrilling or anticipated, such as a special event, achievement, or adventure. It often involves a desire to engage fully in the exciting experience.
\textit{Example Caption: A joyful crowd at a music festival, hands raised, faces beaming with exhilaration, as colorful confetti rains down on them during a thrilling performance.}" \\
\hline
Amusement & "A light-hearted emotion associated with joy and laughter. It arises when something is funny, entertaining, or amusing. It often involves a response to humor, jokes, or playful situations.
\textit{Example Caption: Friends gathered around a table, laughing uncontrollably as they played a hilarious board game, with one person wearing a funny costume and others doubled over in laughter.}" \\
\hline
Awe & "Experienced when encountering something vast, magnificent, or transcendent. It involves a sense of wonder, reverence, and humility in the face of something that is awe-inspiring. Awe can be triggered by natural phenomena like a breathtaking landscape, the night sky, or by human achievements that evoke a sense of grandeur and beauty.
\textit{Example Caption: A breathtaking sunset over a majestic mountain range, casting vibrant hues of orange and purple across the sky, leaving a lone observer standing in awe of nature's grandeur.}" \\
\hline
\bottomrule
\end{tabular}
\caption{Description of emotion labels}
\label{tab:emotions}
\end{table*}

\subsection{Qualitative Samples}
\begin{figure}
    \centering
    \includegraphics[width=1\textwidth]{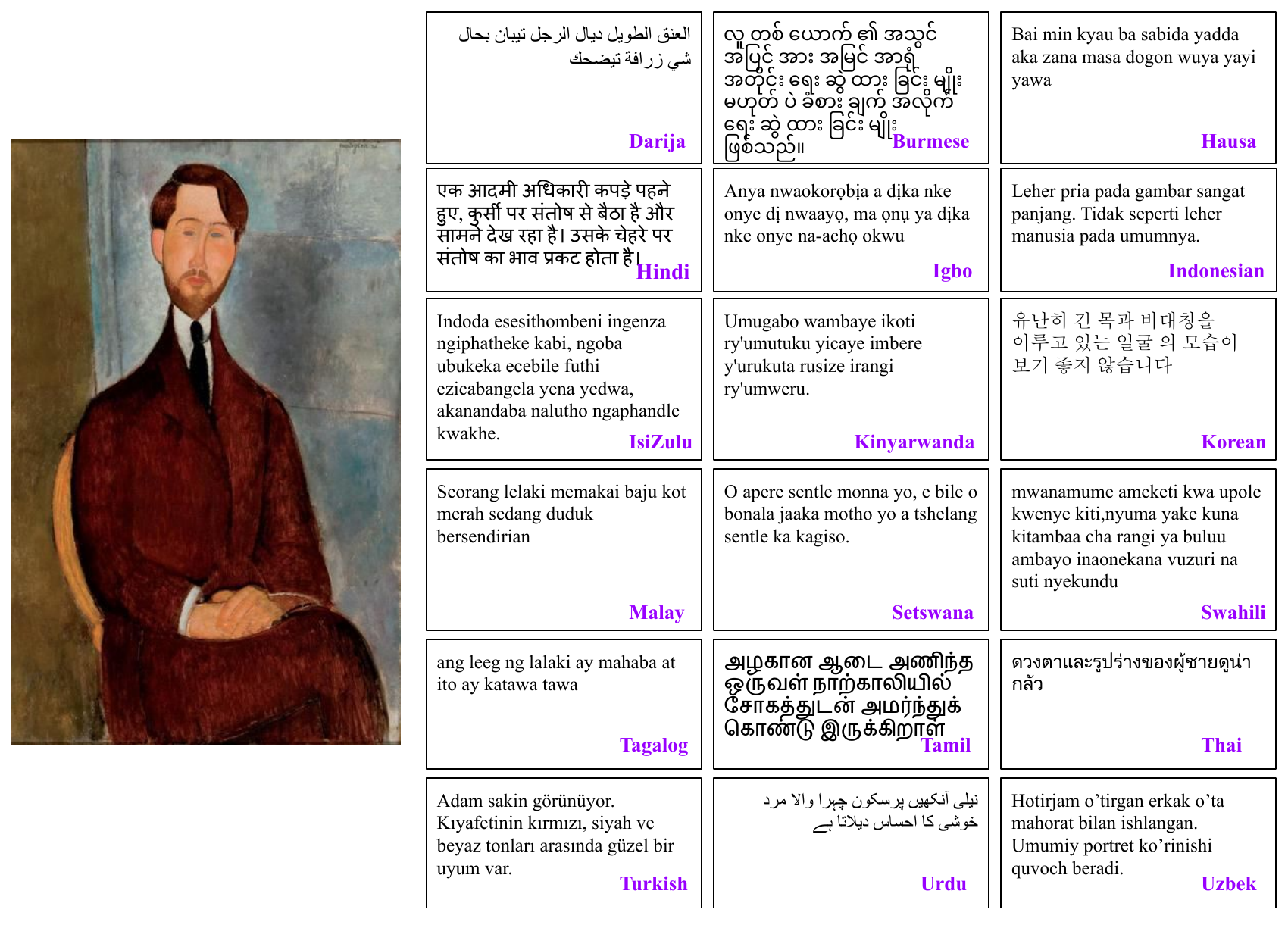}
    \caption{An annotation from ArtELingo-23. Notice the diverse expressions and different points of view across languages.}
    \label{fig:qualitative}
\end{figure}

We show different annotations from ArtELingo-23 in Figure \ref{fig:qualitative}. More examples can be found in our repository.

\subsection{Ethical Concerns}
We received approval for the data collection from KAUST Institutional Review Board (IRB). The IRB requires informed consent; in addition, there are terms of service in AMT. We respected fair treatment concerns from EMNLP (compensation) and IRB (privacy). 

The workers were given full-text instructions on how to complete tasks, including examples of approved and rejected annotations. Participants’ approvals were obtained ahead of participation. Due to privacy concerns from IRB, comprehensive demographic information could not be obtained.

We compensated all annotators with \$0.1 per annotation making the total \$200. The time taken per annotation is on average 45 seconds making the hourly payment \$8 / hour which is above the minimum wage in all of the respective countries. For the language coordinators, we compensated them with \$200 for their reviewing and communication efforts. The workload was lower for the coordinators per annotation. For the head coordinators (hiring and managing multiple language coordinators), they were included as co-authors in this paper.

\clearpage
\section{Evaluation}
\subsection{CLIPScore}
We attempted to use CLIPscore \cite{hessel2021clipscore} to evaluate the generations from our trained model. Since CLIPScore is defined for English language only, we used NLLB \cite{costa2022no} 1.3B model to translate all of our generations into English then evaluated them using CLIPScore. Table \ref{tab:clipscore} reports the scores for InstructBLIP, MiniGPT-4, and ClipCap. According to the results ClipCap is the best performing model. However, upon qualitative inspection of the model output, we observed the very low quality of CLIPCap captions where the generated text is a repetition of the same word. We also observed that the translation model introduce new words that are not faithful to the original translation which might explain the difference in scores (Such analysis was performed on Arabic language only). Accordingly, we decided not to include the results in the main paper.
\begin{table}[h]
    \centering
    \begin{tabular}{l|cc}
    \toprule
       Model  & CLIPScore & RefCLIPScore \\
       \hline
       InstructBlip  & 0.5879 & 0.6566 \\
       ClipCap  & \textbf{0.6111} & \textbf{0.6578} \\
       MiniGPT-4  & 0.4791 & 0.5648 \\
       \hline
       \bottomrule
    \end{tabular}
    \caption{\textbf{CLIPScores and RefCLIPScores.} Zero-Shot performance evaluate using CLIPScore.}
    \label{tab:clipscore}
\end{table}

\subsection{GPT Score}
We utilized GPT-4o-mini to evaluate the quality of the generated captions. GPT4 has been shown to align with human judgement \cite{liu2023g}. Due to cost and rate limitations associated with GPT4 API, we sampled 100 captions from each model in the Zero-Shot performance. We provided the following prompt to GPT-4o-mini:

\begin{lstlisting}
    You will give a score between 1 (worst) and 5 (best) for the following caption and image. Your score will reflect how much the caption is faithful to the image. If the caption is not in {lang} language, give a score of 0. Output only the score without any explanation. The caption is: {caption}
\end{lstlisting}

Table \ref{tab:gptscore} reports the GPTScores. Since the conclusion is similar to the traditional metrics and we only used 100 samples from each model, we decided not to include the results in the main paper to avoid confusion related to sample size.
\begin{table}[h]
    \centering
    \begin{tabular}{l|c}
    \toprule
       Model  & GPTScore \\
       \hline
       InstructBlip  & 0.071 \\
       ClipCap  & 0.61  \\
       MiniGPT-4  & \textbf{0.79}  \\
       \hline
       \bottomrule
    \end{tabular}
    \caption{\textbf{GPTScores.} Zero-Shot performance evaluate using GPTScore.}
    \label{tab:gptscore}
\end{table}

\subsection{Per Language Evaluation}
\label{sec:lang_result}
In this section, we dive deeper into the results reported in the experiments section. We break down the result of each experimental setup by language. We don't report the complete results of CCLM since it didn't generalize to any unseen languages. This shows the importance of Multilingual Large Language Models in creating universal multimodal models that perform well on unseen languages. 

We utilize these results to better understand the limitations of current models. We observe a major bottleneck in the performance of the underlying Multilingual Large Language Model. This is attributed to the lack of powerful counterparts to the English LLMs. We believe that Multilingual LLMs are essential for multilingual multimodal generalization. Hence, we encourage the community to develop more powerful Multilingual LLMs compared to Bloomz.

We face a problem where most metrics' implementation is designed for the English language only. While some implementations can work in other languages where words are separated by white space, this is not generally the case. Especially, for languages like Burmese where the separation between semantic tokens is not as straightforward. As a workaround, we utilize the multilingual tokenizer from the Bloomz model to divide the sentence into tokens. Then, we re-join the produced tokens with whitespaces. We found that these methods work well in most languages.

Nonetheless, such a workaround is not ideal since the tokenizers are usually biased towards languages written in Latin script giving them larger semantic tokens. On the other hand, we found that languages such as Burmese and Vietnamese are tokenized into almost character level. A direct consequence is that the scores scale is drastically different making it impractical to compare performance across different languages. We believe there is a huge room for improvement in multilingual tokenization. A very useful feature would be to guarantee a similar token length across different languages.

In the Few-shot and Seen-Unseen experiments, we used MiniGPt4 since it has the fastest training time as well the best generalization performance. 
\subsubsection{Zero-shot Setting}
Tables \ref{table:zero-shot-bleu}, \ref{table:zero-shot-rouge}, \ref{table:zero-shot-meteor}, \ref{table:zero-shot-cider} report the BLEU, ROUGE, METEOR, and CIDEr scores, respectively. We can immediately notice the poor performance on Burmese and Thai. This reflects the inability of the Bloomz LLM to speak in those languages without finetuning. We can also see that the performance of different models varies greatly in some languages. MiniGPT4 performs the best overall closely followed by clipcap and then InstructBlip.

\begin{table}[ht]
\centering
\scalebox{0.8}{
    \begin{tabular}{c|ccccccccc}
    \toprule
     Model & Burmese & Malay & Korean & Setswana & IsiZulu & Vietnamese & Hindi & Thai & Swahili \\
    \hline
    ClipCap & 0.002 & 0.506 & 0.18 & 0.877 & 0.39 & 6.544 & 4.107 & 0.001 & 1.109  \\
    InstructBlip & 0.001 & 0.449 & 0.028 & 0.572 & 0.356 & 0.673 & 0.097 & 0.0 & 0.58 \\
    MiniGPT4  & 0.002 & 0.337 & 0.979 & 1.223 & 0.635 & 4.21 & 4.344 & 0.0 & 2.097 \\
    \hline
     Model & Kinyarwanda & Urdu & Uzbek & Hausa & Kazakh & Turkish & Darija & Tagalog & Kyrgyz  \\
    \hline
    ClipCap & 0.586 & 2.382 & 0.271 & 0.702 & 0.292 & 0.498 & 0.762 & 0.739 & 0.299 \\
    InstructBlip & 0.401 & 0.351 & 0.257 & 0.713 & 0.315 & 0.663 & 0.002 & 0.737 & 0.407\\
    MiniGPT4  & 1.26 & 2.474 & 0.166 & 0.424 & 0.151 & 0.164 & 0.007 & 0.594 & 0.163 \\
    \hline
     Model & IsiNdebele & Igbo & Emakhuwa & IsiXhosa & Tamil & Yoruba & Indonesian \\
    \hline
    ClipCap & 0.518 & 0.393 & 0.295 & 0.399 & 1.93 & 0.157 & 4.713 \\
    InstructBlip & 0.327 & 0.752 & 0.368 & 0.342 & 0.106 & 0.28 & 0.615 \\
    MiniGPT4  & 0.381 & 1.829 & 0.352 & 0.358 & 2.141 & 0.62 & 2.943 \\
    \bottomrule
    \end{tabular}
}
\caption{Zero-shot Performance on BLEU-4}
\label{table:zero-shot-bleu}
\end{table}

\begin{table}[ht]
\centering
\scalebox{0.8}{
    \begin{tabular}{c|cccccccccc}
    \toprule
     Model & Burmese & Malay & Korean & Setswana & IsiZulu & Vietnamese & Hindi & Thai & Swahili \\
    \hline
    ClipCap & 0.014 & 3.653 & 1.202 & 5.221 & 2.376 & 20.723 & 14.437 & 0.005 & 5.723  \\
    InstructBlip & 0.006 & 1.78 & 0.154 & 3.04 & 2.07 & 2.257 & 0.329 & 0.0 & 2.801 \\
    MiniGPT4  & 0.017 & 1.312 & 2.809 & 5.3 & 2.765 & 12.96 & 13.978 & 0.0 & 7.225   \\
    \hline
     Model & Kinyarwanda & Urdu & Uzbek & Hausa & Kazakh & Turkish & Darija & Tagalog & Kyrgyz  \\
    \hline
    ClipCap & 3.197 & 9.421 & 1.767 & 3.845 & 1.994 & 3.111 & 3.166 & 4.044 & 1.984 \\
    InstructBlip & 2.042 & 1.317 & 1.444 & 3.267 & 1.915 & 3.327 & 0.012 & 3.433 & 1.939 \\
    MiniGPT4  & 5.083 & 9.076 & 0.931 & 1.971 & 0.925 & 0.819 & 0.032 & 2.679 & 0.926 \\
    \hline
     Model & IsiNdebele & Igbo & Emakhuwa & IsiXhosa & Tamil & Yoruba & Indonesian \\
    \hline
    ClipCap & 3.396 & 2.668 & 1.739 & 2.316 & 7.648 & 0.741 & 17.098 \\
    InstructBlip & 1.866 & 3.624 & 1.746 & 1.819 & 0.422 & 1.234 & 2.661 \\
    MiniGPT4  & 1.957 & 6.584 & 1.594 & 1.825 & 7.257 & 2.307 & 10.854 \\
    \bottomrule
    \end{tabular}
}
\caption{Zero-shot Performance on ROUGE}
\label{table:zero-shot-rouge}
\end{table}

\begin{table}[ht]
\centering
\scalebox{0.8}{
    \begin{tabular}{c|cccccccccc}
    \toprule
     Model & Burmese & Malay & Korean & Setswana & IsiZulu & Vietnamese & Hindi & Thai & Swahili \\
    \hline
    ClipCap & 0.054 & 2.122 & 0.989 & 2.208 & 1.111 & 23.151 & 10.699 & 0.316 & 3.307  \\
    InstructBlip & 0.003 & 0.934 & 0.452 & 1.352 & 0.986 & 5.336 & 0.126 & 0.151 & 1.367 \\
    MiniGPT4  & 0.047 & 0.712 & 2.53 & 3.235 & 1.617 & 18.069 & 10.929 & 0.216 & 4.911 \\
    \hline
     Model & Kinyarwanda & Urdu & Uzbek & Hausa & Kazakh & Turkish & Darija & Tagalog & Kyrgyz  \\
    \hline
    ClipCap & 1.439 & 7.312 & 0.854 & 1.707 & 0.827 & 1.36 & 2.029 & 1.798 & 0.917 \\
    InstructBlip & 0.888 & 0.755 & 0.713 & 1.603 & 0.65 & 1.52 & 0.005 & 1.748 & 0.901 \\
    MiniGPT4  & 3.922 & 6.976 & 0.496 & 1.039 & 0.319 & 0.4 & 0.016 & 1.469 & 0.357 \\
    \hline
     Model & IsiNdebele & Igbo & Emakhuwa & IsiXhosa & Tamil & Yoruba & Indonesian \\
    \hline
    ClipCap & 1.456 & 2.063 & 0.831 & 1.025 & 5.239 & 0.415 & 10.429 \\
    InstructBlip & 0.902 & 1.28 & 0.888 & 0.833 & 0.197 & 0.552 & 1.275 \\
    MiniGPT4  & 0.984 & 8.778 & 0.795 & 0.836 & 5.294 & 2.344 & 6.544 \\
    \bottomrule
    \end{tabular}
}
\caption{Zero-shot Performance on METEOR}
\label{table:zero-shot-meteor}
\end{table}

\begin{table}[ht]
\centering
\scalebox{0.8}{
    \begin{tabular}{c|cccccccccc}
    \toprule
     Model & Burmese & Malay & Korean & Setswana & IsiZulu & Vietnamese & Hindi & Thai & Swahili \\
    \hline
    ClipCap & 0.0 & 0.276 & 0.001 & 0.084 & 0.026 & 17.113 & 5.874 & 0.0 & 1.049  \\
    InstructBlip & 0.0 & 0.09 & 0.0 & 0.075 & 0.061 & 2.169 & 0.004 & 0.0 & 0.087 \\
    MiniGPT4  & 0.0 & 0.157 & 0.372 & 0.984 & 0.523 & 13.53 & 5.568 & 0.0 & 3.311 \\
    \hline
     Model & Kinyarwanda & Urdu & Uzbek & Hausa & Kazakh & Turkish & Darija & Tagalog & Kyrgyz  \\
    \hline
    ClipCap & 0.029 & 4.989 & 0.018 & 0.038 & 0.0 & 0.022 & 0.956 & 0.108 & 0.0 \\
    InstructBlip & 0.071 & 0.746 & 0.032 & 0.188 & 0.0 & 0.115 & 0.003 & 0.34 & 0.01 \\
    MiniGPT4  & 2.219 & 5.628 & 0.026 & 0.209 & 0.0 & 0.035 & 0.012 & 0.52 & 0.002 \\
    \hline
     Model & IsiNdebele & Igbo & Emakhuwa & IsiXhosa & Tamil & Yoruba & Indonesian \\
    \hline
    ClipCap & 0.007 & 0.138 & 0.098 & 0.054 & 6.424 & 0.12 & 8.712 \\
    InstructBlip & 0.045 & 0.088 & 0.14 & 0.049 & 0.043 & 0.205 & 0.246 \\
    MiniGPT4  & 0.209 & 1.206 & 0.202 & 0.147 & 6.251 & 0.769 & 5.955 \\
    \bottomrule
    \end{tabular}
}
\caption{Zero-shot Performance on CIDEr}
\label{table:zero-shot-cider}
\end{table}

\clearpage
\subsubsection{Few-shot Setting}
We report the extended results for the One-vs-All Zero-Shot setting in this anonymous file: \url{https://github.com/Mo-youssef/artelingo-28/tree/main/results/minigpt/fewshot.csv}

\subsubsection{One-vs-All Zero-Shot Setting}

We report the extended results for the One-vs-All Zero-Shot setting in this anonymous file: \url{https://github.com/Mo-youssef/artelingo-28/tree/main/results/minigpt/seenunseen.csv}

\subsubsection{Emotion Label Prediction}
\label{sec:emotion_appendix}
We utilize XLM-roBERTa large model from huggingface\footnote{https://huggingface.co/FacebookAI/xlm-roberta-large}. We finetune the model for 5 epochs in all of our experiments. We use AdamW optimizer with $LR=2e-5$ and $eps = 1e-8$ with a linear decay scheduler. We used a batch size of 128 and max caption length of 128 tokens where padding and truncation are utilized to fix the batches number of tokens. The gradients were clipped to have a norm of 1.